%% file: Formatting-Instructions-LaTeX-2023.tex
\documentclass[letterpaper]{article} % DO NOT CHANGE THIS
\usepackage{aaai23}  % DO NOT CHANGE THIS
\usepackage{times}  % DO NOT CHANGE THIS
\usepackage{helvet}  % DO NOT CHANGE THIS
\usepackage{courier}  % DO NOT CHANGE THIS
\usepackage[hyphens]{url}  % DO NOT CHANGE THIS
\usepackage{graphicx} % DO NOT CHANGE THIS
\usepackage{natbib}  % DO NOT CHANGE THIS AND DO NOT ADD ANY OPTIONS TO IT
\usepackage{caption} % DO NOT CHANGE THIS AND DO NOT ADD ANY OPTIONS TO IT
\usepackage{algorithm}
\usepackage{algorithmic}
\usepackage{makecell}
\usepackage{multirow}
\usepackage{amsmath}
\usepackage{amsfonts}
\usepackage{amssymb}
\usepackage{booktabs}
\usepackage{rotating}
\usepackage{xspace}
\usepackage{lineno}
\usepackage{pifont}
\usepackage[table,xcdraw]{xcolor}
\usepackage{tablefootnote}

\newcommand{\cmark}{\ding{51}}%
\usepackage{newfloat}
\usepackage{listings}
\DeclareCaptionStyle{ruled}{labelfont=normalfont,labelsep=colon,strut=off} % DO NOT CHANGE THIS
\floatstyle{ruled}
\newfloat{listing}{tb}{lst}{}
\floatname{listing}{Listing}
\title{PUPS: Point Cloud Unified Panoptic Segmentation}
\author{
    Shihao Su, \equalcontrib\textsuperscript{\rm 1}
    Jianyun Xu, \equalcontrib\textsuperscript{\rm 2}
    Huanyu Wang, \textsuperscript{\rm 1} \\
    Zhenwei Miao, \textsuperscript{\rm 2}
    Xin Zhan, \textsuperscript{\rm 2}
    Dayang Hao, \textsuperscript{\rm 2}
    Xi Li \textsuperscript{\rm 1, 3, 4}\thanks{Corresponding author}
}
\affiliations{
    \textsuperscript{\rm 1}College of Computer Science and Technology, Zhejiang University \\
    \textsuperscript{\rm 2}Alibaba Group \\
    \textsuperscript{\rm 3}Shanghai Institute for Advanced Study, Zhejiang University \\
    \textsuperscript{\rm 4}Shanghai AI Laboratory\\
    shihaocs@zju.edu.cn, xujianyun.xjy@alibaba-inc.com, huanyuhello@zju.edu.cn,\\
    \{zhenwei.mzw, zhanxin.zx\}@alibaba-inc.com, haodayang@gmail.com, xilizju@zju.edu.cn
}

\usepackage{bibentry}
\begin{document}

\maketitle

% \begin{abstract}
% AAAI creates proceedings, working notes, and technical reports directly from electronic source furnished by the authors. To ensure that all papers in the publication have a uniform appearance, authors must adhere to the following instructions.
% \end{abstract}

\newcommand{\ours }[1]{
    \renewcommand{\ours}{#1\xspace}
}
\ours{PUPS}

\input{sections/abstract}

\input{sections/introduction}

\input{sections/related}

\input{sections/method}

\input{sections/experiment}

\input{sections/conclusion}

\newpage
\section*{Acknowledgement}
This work is supported in part by National Key Research and Development Program of China under Grant 2020AAA0107400, Zhejiang Provincial Natural Science Foundation of China under Grant LR19F020004, National Natural Science Foundation of China under Grant U20A20222, National Science Foundation for Distinguished Young Scholars under Grant 62225605, Alibaba-Zhejiang University Joint Research Institute of Frontier Technologies, Ant Group, and sponsored by CAAI-HUAWEI MindSpore Open Fund.

\bibliography{aaai23}

\end{document}

%% file: sections/abstract.tex
\begin{abstract}
    Point cloud panoptic segmentation is a challenging task that seeks a holistic solution for both semantic and instance segmentation to predict groupings of coherent points. 
    Previous approaches treat semantic and instance segmentation as surrogate tasks, and they either use clustering methods or bounding boxes to gather instance groupings with costly computation and hand-crafted designs in the instance segmentation task.
    % Instance segmentation is typically more difficult than semantic segmentation and previous approaches either used clustering methods to post create instances or used detection head to gather proposals with lots of computation and hand-crafted designs.
    % Instance segmentation is typically more complicated than semantic segmentation. Previous approaches split them as two independent branches, and either used cluster methods or detection boxes to gather instances, which employ lots of computation and hand-crafted designs.
    In this paper, we propose a simple but effective \textbf{p}oint cloud \textbf{u}nified  \textbf{p}anoptic \textbf{s}egmentation (\ours) framework, which use a set of point-level classifiers to directly predict semantic and instance groupings in an end-to-end manner.
    % To realize \ours, we adopt three key designs in our method. First and foremost, we introduce bipartite matching to our training pipeline so that our classifiers are able to discriminate between instances, getting rid of hand-crafted post-processing, e.g. anchors and Non-Maximum Suppression (NMS).
    To realize \ours, we introduce bipartite matching to our training pipeline so that our classifiers are able to exclusively predict groupings of instances, getting rid of hand-crafted designs, e.g. anchors and Non-Maximum Suppression (NMS).
    In order to achieve better grouping results, we utilize a transformer decoder to iteratively refine the point classifiers and develop a context-aware CutMix augmentation to overcome the class imbalance problem.
    % Second, in order to achieve better grouping results, we utilize a transformer decoder to iteratively refine the point classifiers.
    % Third, we develop a context-aware CutMix augmentation to overcome the class imbalance problem.
    As a result, \ours achieves \textbf{1st} place on the leader board of SemanticKITTI panoptic segmentation task and state-of-the-art results on nuScenes.
    
\end{abstract}

%% file: sections/introduction.tex
\section{Introduction}
    % As one of the most challenging problems in computer vision, panoptic segmentation~\cite{panopticsegmentation} seeks a holistic solution to both semantic segmentation and instance segmentation. Shortly after the researchers proposed the question in image data, two of the most popular LiDAR datasets for autonomous driving release their panoptic version, extending the research area to point cloud data. 
    As one of the most challenging problems in computer vision, panoptic segmentation~\cite{panopticsegmentation} seeks a holistic solution to both semantic segmentation and instance segmentation. Shortly after the researchers proposed the question in image data, two LiDAR datasets~\cite{SemanticKITTI, nuScenesSeg} for autonomous driving extend the research area to point cloud data. 
    % These emerging challenges reveals that perception system of autonomous vehicles demands understanding of the environment in terms of both semantic and instance since the objective of LiDAR panoptic segmentation is to group LiDAR points into \emph{thing} classes 
    These emerging challenges aim at assigning points with groupings of countable \emph{thing} instances and uncountable \emph{stuff} classes, revealing that perception system of autonomous vehicles demands understanding of the environment in terms of both semantic level and instance level through point cloud sensors.
    
\begin{figure}[t]
        \centering
        \includegraphics[width=0.4\textwidth]{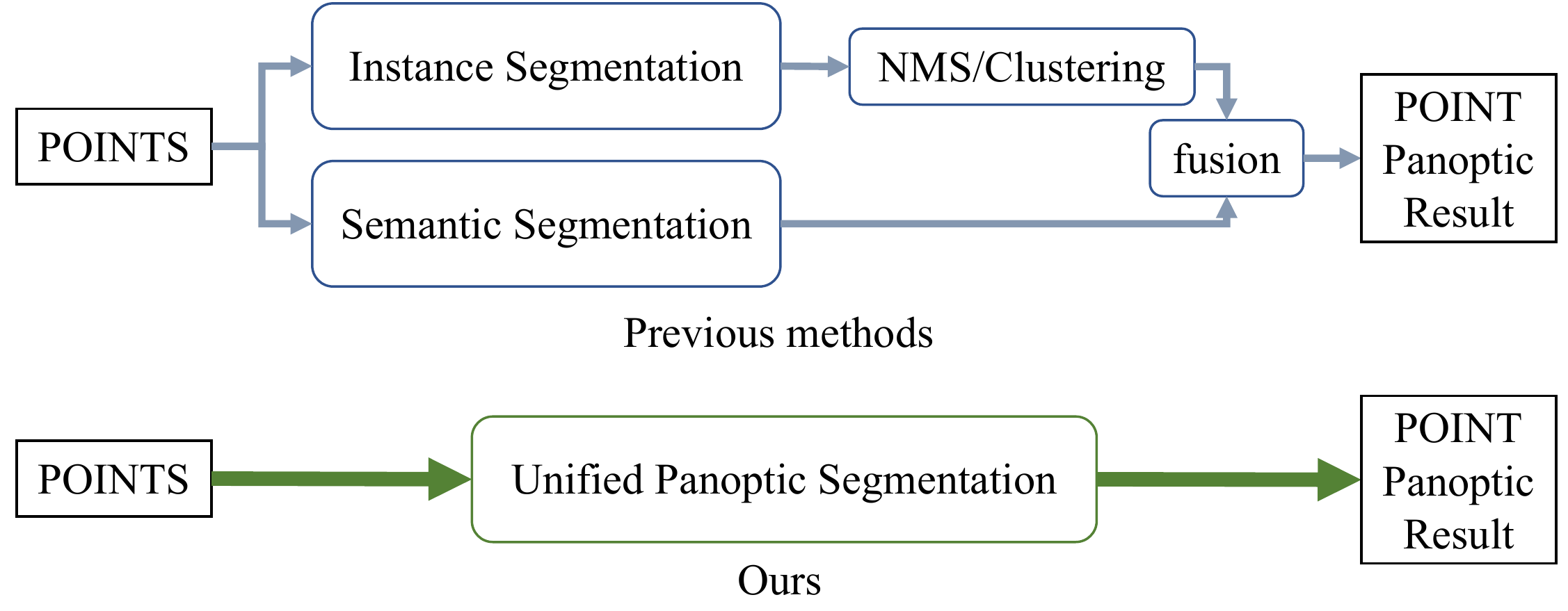}
        \caption{Illustration of previous framework and \ours.}
        % \caption{Upper: pipeline of previous methods. Instance and semantic are separately handled. They employ hand-crafted or costly computation to group instances in their instance branch and fuse instance predictions and semantic predictions to eventually get panoptic results. Lower: \ours directly predict result of panoptic segmentation without post-processing or hand-crafted design.}
        \label{fig:intro}
    \end{figure}

    To solve point cloud panoptic segmentation, previous efforts can be divided into two streams: proposal-based methods and proposal-free methods. As the name indicates, proposal-based methods rely on proposals generated by an object detection head to get instance segmentation
    % so that region of interest (RoI) is cropped for instance segmentation branch 
    and employ an extra semantic branch for semantic segmentation. Besides their cascaded structure, this stream of methods involves lots of hand-crafted components such as proposals and non-maximum suppression (NMS). As for proposal-free methods, they introduce clustering-based methods in their instance branch based on the predicted offsets to instance centers. Similarly, an extra semantic branch is attached for semantic segmentation. 
    % Requiring careful tuning, the clustering algorithms are also called post-processing in other methods. 
    Although outstanding results have been achieved by these methods on different benchmarks, there are two main drawbacks as shown in the upper part of Figure~\ref{fig:intro}: 1) they treat semantic and instance segmentation as surrogate tasks, which does not truly solve panoptic segmentation holistically; 2) their instance branch involves many hand-crafted components and post-processing, which is complicated and time-consuming.

    % Moreover, the performance of multi-branch framework depends one of the branches
    
    % simple effective unified end-to-end mask bipartite transformer
    
    % Inspired by recent developments in image segmentation~\cite{cheng2021maskformer, max-deeplab, panoptic_segformer, zhang2021knet}, we propose \ours, a simple but effective \textbf{p}oint cloud \textbf{u}nified  \textbf{p}anoptic \textbf{s}egmentation framework with no hand-crafted design or post-processing as shown in the lower part of Figure~\ref{fig:intro}. In essence, the aim of point cloud panoptic segmentation is to predict groupings of coherent points. Our method unifies point cloud instance and semantic segmentation by introducing a set of point-level binary classifiers to learn and directly predict these groupings in an end-to-end manner. 
    % To assign semantic classes to the groupings, we perform a semantic classification subjected to the classifiers and their groupings. Afterwards, the output of point cloud panoptic segmentation is obtained simultaneously in this simple pipeline. 

    % To realize \ours, we allocate a fixed number of point-level binary classifiers and introduce bipartite matching to our training pipeline. 
    % In order for the fixed number of classifiers to learn from varying number of ground-truth groupings in different samples, a mapping from the predictions to ground-truth is required. 
    % Bipartite matching is particularly suitable for this case because one-to-one mapping enables our classifiers to learn from unique instances, thus enabling our model to exclusively predict groupings without post-processing.  
    
    Inspired by recent developments in image segmentation~\cite{cheng2021maskformer, max-deeplab, panoptic_segformer, zhang2021knet}, we propose \ours, a simple but effective \textbf{p}oint cloud \textbf{u}nified  \textbf{p}anoptic \textbf{s}egmentation framework to solve the challenges above. In essence, the aim of point cloud panoptic segmentation is to predict groupings of coherent points. \ours unifies point cloud instance and semantic segmentation as a classifier-assigning problem. More specifically, \ours allocates a set of point-level classifiers, learning to assign them to exclusive instances or semantic classes. By utilizing bipartite matching in the training phase, we achieve \ours in an end-to-end manner and it is able to predict exclusive groupings with no hand-crafted design or post-processing as shown in Figure~\ref{fig:intro}.

    % Although the number of groupings in a single scan of point cloud varies, the training process of our method is still conducted in an end-to-end manner. 
    % Note that there is no proposal designing, NMS or clustering involved in the pipeline.
    
    % To realize \ours, we introduce bipartite matching to our training pipeline so that our model can exclusively predict the groupings without post-processing. \ours allocates a fixed number of point-level binary classifiers which is usually not equal to the number of ground-truth groupings in a single scan of point cloud. 
    % In order for the fixed number of classifiers to learn from varying number of ground-truth groupings, a mapping from the predictions to ground-truth is required. 
    % Bipartite matching is particularly suitable for this case because one-to-one mapping enables our classifiers to discriminate between instances. Point cloud groupings can be directly predicted with the classifiers.
    
    % transformer decoder
    In addition to predicting exclusive groupings, we adopt two designs to produce better grouping results.
    First, we utilize a transformer decoder to refine the classifiers. In each stages of refinement, our point-level classifiers query point features from backbone and generate refined classifiers. Afterwards, the classifiers integrate the feature of corresponding instances and semantics, enhancing their ability to distinguish between groupings. Moreover, we employ a classifier self-attention to incorporate global relations into the classifiers. After the classifiers are refined, new point groupings are produced and can be further used to refine the classifiers in the next stage together with the refined classifiers.
    Second, in order to alleviate class imbalance and train the classifiers more sufficiently, we design a context-aware CutMix~\cite{SECOND, RPVNet, Panoptic_PHNet} augmentation. We cut instances from training scans and mix them with the instances in the current scan based on their background to avoid damaging their context so that performance is improved in return.
    % It hampers the learning of our point-level classifiers on minor class instances since the number of instances varies greatly among classes. 
    % However, cutting instances in training scans and mix them in the current scan randomly spoils the context of instances. Therefore, we mix them based on their background to avoid damaging their context so that performance on segmentation is improved.
    
    To evaluate the effectiveness of our proposals, we conduct extensive experiments on two point cloud panoptic segmentation datasets. Our method ranks 1st on the leader board of SemanticKITTI~\cite{SemanticKITTI} 
    % \footnote{Screenshot of the leader board is included in supplementary material}
    and achieves state-of-the-art results on nuScenes~\cite{nuScenes}.
    
    To sum up, the contributions of this paper are listed below:
    \begin{itemize}
        % \item To the best of our knowledge, we are the first to introduce bipartite matching in the training of a LiDAR point cloud model. Our pipeline provides a solution to LiDAR point cloud panoptic segmentation that is not dependent on proposals, NMS or post-processing.
        % \item To the best of our knowledge, we are the first to propose an end-to-end panoptic segmentation method to unify instance and semantic segmentation for point cloud data. Our method directly predict the \emph{thing} and \emph{stuff} groupings with a simple pipeline. 
        \item To the best of our knowledge, \ours is the first simple but effective \textbf{p}oint cloud \textbf{u}nified \textbf{p}anoptic \textbf{s}egmentation framework, using a set of point-level classifiers to directly predict semantic and instance groupings.
        \item To get rid of post-processing and hand-crafted designs, we introduce bipartite matching in our training so that the classifiers are able to exclusively predict groupings.
        \item We utilize a transformer decoder to iteratively refine the classifiers with point features to produce more accurate groupings of points.
        \item To encounter class imbalance, we adopt a context-aware CutMix strategy to enhance the performance of segmentation by preserving the context of instances.
        \item We achieve rank 1 performance on the leader board of SemanticKITTI panoptic segmentation task and SOTA results on nuScenes.
    \end{itemize}

%% file: sections/related.tex
\section{Related Work}
Panoptic Segmentation aims to divide an input sample into countable \emph{thing} instances or uncountable \emph{stuff} classes. The output of panoptic segmentation is to assign element-wise label with both instance ID and semantic class. For input modal of LiDAR point cloud or image, panoptic segmentation models follow two typical frameworks: proposal-based and proposal-free.
\subsection{LiDAR Point Cloud Panoptic Segmentation}
    \subsubsection{Proposal-based methods} This kind of methods are usually formulated in a two-stage manner: segmentation after detection~\cite{LPSAD, MOPT}. Moreover, SemanticKITTI~\cite{SemanticKITTI} and nuScenes~\cite{nuScenes} report results by joining state-of-the-art point cloud object detection methods and point cloud semantic segmentation methods. Taking in range-view images, EfficientLPS~\cite{EfficientLPS} utilizes a instance branch to predict classes, bounding boxes and masks for \emph{thing} classes and fuse semantic feature to predict \emph{stuff} classes. After post-processing, the range-view result is projected back to point-wise result. It is worth noting that hand-crafted components such as anchors or NMS are often involved in these methods. 
    \subsubsection{Proposal-free methods} As for proposal-free ones, they usually predict instance centers and point-wise offset to centers to output panoptic segmentation result~\cite{Panoptic-PolarNet, DS_Net}. Recently, Panoptic-PHNet~\cite{Panoptic_PHNet} introduces a K-NN transformer to predict more accurate offsets. Additionally, proposal-free methods~\cite{Panoster, DS_Net, Panoptic_PHNet} often involve clustering algorithms to cluster points to instances. GP-S3Net~\cite{GP-S3Net} propose a novel graph-based clustering method to effectively predict instances from over-segmented clusters. 
    
    \ours directly groups point cloud without any bounding box proposals, hand-crafted post-processing or clustering algorithms. It is worth noting that these hand-crafted components in both proposal-based/free methods require lots of computation and careful tuning.
\subsection{Image Panoptic Segmentation}
    \subsubsection{Proposal-based methods} This stream of methods follow the pipeline that bounding boxes are first obtained and masks of each bounding box are predicted afterwards, such as Panoptic-FPN~\cite{Panoptic-FPN}. These methods fuse their masks of \emph{thing} classes and masks of \emph{stuff} classes with merging modules~\cite{Liu_2019_CVPR, Li_2019_CVPR, Porzi_2019_CVPR}.
    \subsubsection{Proposal-free methods} Proposal-free methods solves panoptic segmentation by employing two separate branches to predict semantic masks and group pixels to instances. One of the most popular grouping methods is instance center regression, which predicts pixel-level offsets to instance centers~\cite{Neven_2019_CVPR, cheng2020panoptic}. Recently, following DETR~\cite{DETR}, multiple works introduce bipartite matching in their training~\cite{zhang2021knet, max-deeplab, cheng2021maskformer, li2022maskdino}, simplifying the process of panoptic segmentation.
    
    Inspire by the methods that introduce bipartite matching into their training pipeline, we propose \ours, the first framework on point cloud data which is able to exclusive predict panoptic groupings of points and can be trained in an end-to-end manner.

%% file: sections/method.tex
\section{Method}
\begin{figure*}[t]
    \centering
    \includegraphics[width=\textwidth]{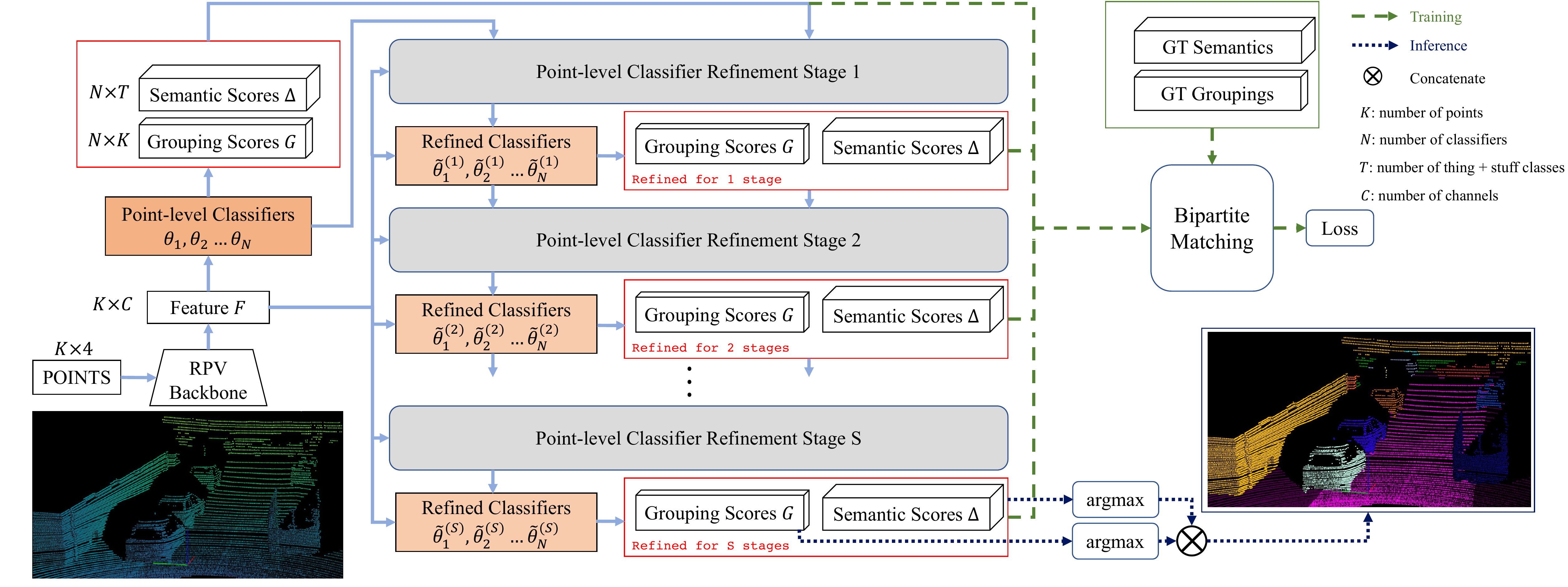}
    \caption{Pipeline of \ours. Point features are first encoded by a RPV backbone~\cite{RPVNet} and fed into the unrefined classifiers to get initial groupings and semantics. Then, point features activated by corresponding groupings are integrated into the classifiers and a self-attention is applied to the classifiers to produce refined classifiers. For simplicity, we omit the superscripts of classifiers in Section~\ref{sec:classifier_refine}. With the refined classifiers, more accurate groupings and semantics are obtained. To clarify, groupings and semantics of all stages will be supervised by ground-truth with bipartite matching in training and only the groupings and semantics of the last stage will be used to output segmentation results in inference.}
    \label{fig:pipeline}
\end{figure*}

% In this section, we first present the definition of point cloud panoptic segmentation. Second, we present the  network architecture of \ours and shows how our framework is able to predict panoptic results without post-processing. Third, we elucidate the bipartite matching training scheme which enable the classifiers to exclusively predict point cloud instances. Fourth, we elaborate on the classifier refinement based on transformer decoder and enhance the grouping results. Lastly, a context-aware instance sampling augmentation is devised to enhance the classifiers by preserving the context of the instances. 

In this section, we first state the definition of point cloud panoptic segmentation in Section~\ref{sec:problem_formulation}. Then, we present the network architecture of \ours in Section~\ref{sec:pups}, along with its two core components, i.e., bipartite matching (Section~\ref{sec:bipartite_matching}) and classifier refinement (Section~\ref{sec:classifier_refine}). Lastly, we show a context-aware CutMix for instances in Section~\ref{sec:cutmix}.

% In this section, we first present the definition of point cloud panoptic segmentation and the network architecture of \ours. Second, we elaborate on the multi-stage refining scheme and explain the procedure of the sample-adaptive classifier generation. Third, we elucidate the bipartite matching training scheme which enable the classifiers to distinguish point cloud instances. Lastly, a context-aware instance sampling augmentation is devise to enhance the classifiers by preserving the context of the points. 

\subsection{Problem Formulation}
\label{sec:problem_formulation}
Point cloud panoptic segmentation aims at grouping a point cloud $P \in \mathbb{R}^{K\times 4}$ of $K$ points into a set of \emph{thing} instances and \emph{stuff} classes, among which \emph{thing} instances refer to countable objects (e.g. person, car, bicycle) and \emph{stuff} classes refers to uncountable backgrounds (e.g. road, terrain, vegetation). As shown in Equation~\ref{eq:problem_formulation}, ground-truth groupings in a point cloud are defined as:
\begin{equation}
    \label{eq:problem_formulation}
    \{y_i\}_{i=1}^M = \{(g_i, c_i)\}_{i=1}^M,
\end{equation}
where $g_i \in \{0,1\}^K$ is a ground truth binary mask indicating which points belong to group $i$, $c_i$ is the semantic class of group $i$, and $M$ is the number of ground truth groupings in the point cloud. 
% Let $C_{\text{thing}}$ be the number of \emph{thing} classes and $C_{\text{stuff}}$ be the number of \emph{stuff} classes.
Note that the groupings are mutually exclusive, i.e.
% , $\forall i \neq j, g_i \cap g_j = \emptyset$
, each point in a point cloud belongs to either an instance of \emph{things} or a background \emph{stuff}. In this way, each point is assigned to a group ID and a semantic class.

\subsection{Point Cloud Unified Panoptic Segmentation}
\label{sec:pups}
% Given the definition in Section~\ref{sec:problem_formulation}, we allocate $N$ learnable point-level classifiers to predict the groupings for both distinct \emph{thing} instances and background \emph{stuff} in a unified manner. We denote the learnable parameters of the classifiers as $\theta = \{\theta_i\}_{i=1}^N$. As shown in the inference pipeline of Figure~\ref{fig:pipeline}, with the point-level classifiers, it is simple to obtain grouping scores $G = \{\hat{g}_i\}_{i=1}^N, \hat{g}_i \in [0,1]^K$ for each point with respect to each classifier. And $\hat{g}_i$ is computed as followed. 

Given the definition in Section~\ref{sec:problem_formulation}, we allocate $N$ learnable point-level classifiers to predict the groupings for both distinct \emph{thing} instances and background \emph{stuff} in a unified manner. We denote the learnable parameters of the classifiers as $\theta = \{\theta_i \;|\; \theta_i \in \mathbb{R}^C \}_{i=1}^N$, where $C$ is the number of channels. As shown in the inference pipeline of Figure~\ref{fig:pipeline}, $K$ points are fed into a backbone to extract point-wise feature $F$. Using the feature $F$ and the parameters $\theta$, \ours generate two vital scores for panoptic segmentation results: grouping scores $G = \{\hat{g}_i \;|\; \hat{g}_i \in [0,1]^K\}_{i=1}^N$ and semantic scores $\Delta = \{\Delta_i \;|\; \Delta_i \in \mathbb{R}^{T}\}_{i=1}^N$, where $T$ is the number of \emph{thing} and \emph{stuff} classes. 

First of all, grouping scores $G$ indicate the probability of the $K$ points belonging to $N$ groups, which are used to decide the group ID for each point. Similarly, semantic scores $\Delta$ indicates the probability of the $N$ groupings belonging to $T$ semantic classes, which are used to assign semantic classes to groupings, and further decide semantic classes for each point.

% Specifically, with the parameters $\theta$ of the point-level classifiers, it is simple to obtain grouping scores $G = \{\hat{g}_i|\hat{g}_i \in [0,1]^K\}_{i=1}^N$ for each point with respect to each classifier: 

Specifically, with the parameters $\theta$ of classifiers and the feature $F$ of points, we utilize a simple matrix multiplication and a sigmoid, denoted as $\delta(\cdot, \cdot)$, to obtain grouping scores $G$ for each point with respect to each classifier:

% Given the definition in Section~\ref{sec:problem_formulation}, we allocate $N$ learnable point-level classifiers and assign them to predict the groupings for both distinct \emph{thing} instances and background \emph{stuff} in a unified manner. We denote the learnable parameters of the classifiers as $\theta = \{\theta_i\}_{i=1}^N$. As shown in the inference pipeline of Figure~\ref{fig:pipeline}, with the point-level classifiers, it is simple to obtain grouping scores $G = \{\hat{g}_i|\hat{g}_i \in [0,1]^K\}_{i=1}^N$ for each point with respect to each classifier. And $\hat{g}_i$ is computed by feeding the point features into the classifiers as followed: 
\begin{equation}
    \label{eq:grouping_scores}
    \hat{g}_i = \delta(\theta_i, F), i=1,\hdots,N,
\end{equation}
where $F \in \mathbb{R}^{K \times C}$ is the point-level features of $K$ points. 
% output by a backbone and $\delta(\theta_i, F)$ is a function that performs matrix multiplication between $\theta_i$ and a sigmoid afterwards. 
% It is easy to find out that $\hat{g}_i$ is a soft version of $g_i$, indicating the probability of a point being in the group.

Similarly, we utilize $\delta(\cdot, \cdot)$ to predict a semantic score $\Delta_i \in \mathbb{R}^{T}$ for each point-level classifier, where $T$ is the number of \emph{thing} and \emph{stuff} classes:
\begin{equation}
    \label{eq:semantic_scores}
    \Delta_i = \delta(\psi, \theta_i), i=1,\hdots,N.
\end{equation}
To clarify, $\psi \in \mathbb{R}^C$ stands for a set of learnable parameters. 
% $\delta(\psi, \theta_i)$ performs matrix multiplication between two sets of learnable parameters $\psi$ and $\theta_i$ and a sigmoid afterwards. 
% $\Delta \in \mathbb{R}^{T}$, where $T$ is the number of \emph{thing} and \emph{stuff} classes.
% both $G$ and the parameters of the classifiers to predict a semantic score simplex $\Delta \in \mathbb{R}^{T}$, where $T$ is the number of \emph{thing} and \emph{stuff} classes. 
% To this end, the scores output by \ours can be organized in Equation~\ref{eq:output_scores}.
% % in Equation~\ref{eq:grouping_scores},
% % \begin{equation}
% %     \label{eq:grouping_scores}
% %     G = \{\hat{g}_i\}_{i=1}^N
% % \end{equation}
% \begin{equation}
%     \label{eq:output_scores}
%     \{\hat{s}_i\}_{i=1}^N = \{(\hat{g}_i, \Delta_i)\}_{i=1}^N.
% \end{equation}
% By comparing Equation~\ref{eq:problem_formulation} and Equation~\ref{eq:output_scores}, it is easy to find out that

To output the result of panoptic segmentation, PUPS assigns semantic class $c_i$ to $\hat{g}_i$ and groupings to points by:
% Then, the semantic scores of the assigned class is used to modulate the grouping scores as follow,
% \begin{equation}
%     \Tilde{g}_i = \hat{g}_i \cdot \max (\Delta_i).
% \end{equation}
\begin{gather}
    \label{eq:output}
    \hat{c}_i = \arg \max \Delta_i, \\
    \hat{z}_{i,k} = 
        \begin{cases}
            1 & \text{if} \quad \hat{g}_{i,k} = \underset{i}{\max} \, (\hat{g}_{i,k}) \\
            0 & \text{otherwise}
        \end{cases},\\
    \{\hat{y}_i\}_{i=1}^N = \{(\hat{z}_i, \hat{c}_i)\}_{i=1}^N,
\end{gather}
Our prediction is organized similarly to Equation~\ref{eq:problem_formulation} and group ID and semantic class for each point is obtained.

% To this end, \ours is able to predict the groupings directly. However, possibilities are that some classifiers predict the same instances, resulting in over-segmented instance since there is no post-processing to remove duplicate predictions.

Now, \ours is able to predict the groupings directly. Instead of adding post-processing at test time, we use bipartite matching (Section~\ref{sec:bipartite_matching}) in our training pipeline to prevent multiple classifiers from predicting the same instance.

\subsection{Bipartite Matching}
\label{sec:bipartite_matching}
% Our proposed method is able to avoid over-segmenting instances with duplicate prediction and exclusively predict the grouping if the point-level classifiers' ground truth grouping is exclusive with respect to each other. Otherwise, different classifiers learns to predict the same grouping, thus giving rise to duplicate prediction. 
One solution to the aforementioned problem is to make our point-level classifiers learn from exclusive ground truth groupings.
% The definition of point cloud panoptic segmentation states that the number of instances in a scan of point cloud varies, i.e., the number of ground-truth groupings $M$ in Equation~\ref{eq:problem_formulation} is different in different samples. A fixed mapping from the  classifiers to the instances is not appropriate. 
Therefore, a one-to-one mapping from the $M$ ground-truth groupings to the $N$ classifiers is needed.
% to train them in an end-to-end manner.
Inspired by recent application of bipartite matching in object detection~\cite{DETR} and image segmentation~\cite{cheng2021maskformer, panoptic_segformer, zhang2021knet}, bipartite matching is able to assign one ground-truth to only one prediction according to a cost matrix. This one-to-one rule plays a vital role in exclusively predicting the point cloud panoptic segmentation results for the reason that there are no classifier assigned to learn the same \emph{thing} instance or \emph{stuff} class, reducing the possibility of duplicate predictions. It also prevents the classifiers from only focusing on easy groupings because all ground truths are mapped, reducing the bias of the model. Since instance ID prediction is not required for \emph{stuff} classes, several classifiers are constantly mapped to the ground truth of \emph{stuff} classes.
\subsubsection{Cost Computation} To match the predictions $\{\hat{s}_i\}_{i=1}^N=\{(\hat{g}_i, \Delta_i)\}_{i=1}^N$ and ground-truths $\{y_i\}_{i=1}^M\{(g_i, c_i)\}_{i=1}^M$, we compute the cost matrix based on their pairwise accordance in terms of both points and groups. For simplicity, we term the match cost and the training loss between prediction and ground-truth with the same notation:
\begin{equation}
    \label{eq:match_loss}
    \mathcal{L}_{\text{match}} = \alpha \mathcal{L}_{\text{dice}} +  \beta \mathcal{L}_{\text{focal}} + \gamma \mathcal{L}_{\text{CE}},
\end{equation}
% where dice loss $\mathcal{L}_{\text{dice}}$~\cite{diceloss} and cross entropy loss ($\mathcal{L}_{\text{CE}}$) are for point-wise accordance between $\forall 1 \le i \le N, 1 \le j \le M, \hat{g}_i, g_j$, and focal loss~\cite{2017Focal} is for group classification between $\forall 1 \le i \le N, 1 \le j \le M, \Delta_i, c_j$. For classifiers that are not assigned with any ground truth, they are masked as negative.
where dice loss $\mathcal{L}_{\text{dice}}$~\cite{diceloss} and cross entropy loss ($\mathcal{L}_{\text{CE}}$) are for point-wise accordance between $\hat{g}_i$ and $g_j$ ($\forall 1 \le i \le N, 1 \le j \le M$). The focal loss~\cite{2017Focal} is for group classification between $\Delta_i$ and $c_j$ ($\forall 1 \le i \le N, 1 \le j \le M$). For classifiers that are not assigned with any ground truth, they are masked as negative.

% during the training of this point cloud.

In summary, Section~\ref{sec:pups} resolves the problem of predicting the groupings and Section~\ref{sec:bipartite_matching} enables the classifiers to learn from exclusive ground truth so that there is no need to apply post-processing to remove duplicate predictions or clustering algorithms to coalesce segmented groupings.

\subsection{Classifier Refinement}
\label{sec:classifier_refine}
Although the process is simple as stated in Section~\ref{sec:pups}, it is challenging to accurately classify the unordered points into groups. Being different from classifying points into semantic classes, panoptic segmentation demands a further step to discriminate instance information within one semantic class. Moreover, the predicting process is purely point-based, i.e., points are treated as isolated only with implicit spatial information, and appearance encoded in their backbone features. 
% Thus, chances are that the classification process introduces noise from other groupings.
Thus, classifying the points into groups only once may introduce noises from other groups

% Therefore, 
To fulfill the aforementioned demand and overcome the problem, we employ a transformer decoder with $S$ stages to refine the point-level classifiers with point features as shown in Figure~\ref{fig:pipeline}. 
% In each stage of the transformer decoder, the process of refinement are divided into three part: point feature query, refinement with instance and semantic feature, and classifier self-attention.
% The classifier refinement first gathers instance and semantic features around an instance. Then, the features are integrated into the classifiers. Afterwards, the context of among the classifiers are modeled with a multi-head self-attention.
In each stage of the transformer decoder, the process of refinement are divided into three parts: 1) classifier feature query. It gathers point features for each classifier. 2) classifier update. It updates the classifiers with the gathered features. 3) classifier self-attention. It further models the context information between classifiers.

% The classifier refinement first highlight activated points and suppress noisy ones with the grouping score $\hat{g}_i$.
% % , building on the assumption that the classifier before refinement is able to roughly group points of an instance. 
% Afterwards, the local feature around an instance is aggregated, containing instance and semantic information. Moreover, the context of the classifier is modeled with self-attention. As a result, the refinement process utilize both local feature queried by the classifiers and the context information between classifiers.

\subsubsection{Classifier Feature Query}
% First, given the point feature $F\in \mathbb{R}^{K\times C}$ and the parameters of our point-level classifiers, we query the point features with the point-level classifiers and obtain grouping score $\hat{g}_i \in [0,1]^K$ for classifier $i$. The grouping score serves as an attention map for the $i$th classifier with respect to every single point in the point cloud so that the most related features are collected as instance and semantic information.
First, we query the point features with grouping scores from Equation~\ref{eq:grouping_scores} using the parameters of the point-level classifiers fed into this stage. The grouping score serves as an attention map for the $i$th classifier with respect to every single point in the point cloud so that the most related features are collected as instance and semantic information.
% The process of point feature query follows the computation in Equation~\ref{eq:grouping_scores}. 
% Specifically, $\delta_1$ contains a fully-connected layer followed by a layer normalization and a sigmoid. $\hat{g}_i \in [0,1]^K$ serves as an attention map for the $i$th classifier with respect to every single point in the point cloud. 
With the point features and their attention, the discriminative instance and semantic feature $F^{\theta_i} \in \mathbb{R}^{C}$ of the grouping $i$ is obtained by:
\begin{equation}
    \label{eq:aggregate_features}
    F_{\theta_i} = \frac{1}{K} \sum_{k=1}^K \hat{g}_{i,k} \cdot F_k.
\end{equation}
For simplicity, we omit the superscript used in Figrue~\ref{fig:pipeline} and  $\theta = \{\theta_i\}_{i=1}^N$ stands for the parameters of the classifiers fed into this stage.

\subsubsection{Classifier Update}
Since the objective of classifiers is to learn the distinct groupings of points, we integrate the discriminative instance and semantic features into the parameters of the classifiers so that they are able to retrieve instance points that are missing in the current groups and rule out noisy ones, thus producing more accurate results. Specifically, we first project the features into the space of classifiers' parameter and employ a learnable momentum $m$ to control the extent of integration. The calculation of the projection and momentum $m$ is as followed:
% With obtained feature $F_{\theta_i}$ in Equation~\ref{eq:aggregate_features}, we incorporate the instance and semantic information into the classifiers by first projecting it to the space of classifiers' parameters and modulate the parameters by:
% \begin{equation}
%     \label{eq:classifier_feature}
%     \Tilde{F}_{\theta_i} = \varphi_1(F_{\theta_i}) \odot \theta_i,
% \end{equation}
% where $\Tilde{F}_{\theta_i} \in \mathbb{R}^{C}$ is the combined features of classifiers parameters, instance information and semantic information, $\varphi_1$ is a linear transformation, and $\odot$ is element-wise product. 
% To obtain the refined classifier, we learn a momentum factor $m \in \mathbb{R}$ to gradually update the parameters of the classifier by:
% \begin{align}
%     \label{eq:refine_momentum}
%     m = 1 & - \sigma (\varphi_1(\Tilde{F}_{\theta_i})) \\
%     \Tilde{\theta}_i = (1-m) &\cdot \varphi_2(\Tilde{F}_{\theta_i}) + m \cdot \theta_i,
% \end{align}
\begin{align}
    \label{eq:refine_momentum}
    m = 1 & - \sigma (\varphi_1({F}_{\theta_i})) \\
    \Tilde{\theta}_i = (1-m) &\cdot \varphi_2({F}_{\theta_i}) + m \cdot \theta_i,
\end{align}
where $\sigma$ is a non-linear function sigmoid and $\varphi_1$, $\varphi_2$ are linear transformations.

% Since the objective of classifiers is to learn the distinct groupings of the points, we integrate the activated point features into the parameters of the classifiers so that they are able to refine the grouping and produce more accurate results.
% With obtained feature $F_{\theta_i}$ in Equation~\ref{eq:aggregate_features}, we incorporate the instance and semantic information into the classifiers by projecting them to the same space by:
% \begin{equation}
%     \label{eq:classifier_feature}
%     \Tilde{F}_{\theta_i} = \varphi_1(F_{\theta_i}) \odot \varphi_2(\theta_i),
% \end{equation}
% where $\Tilde{F}_{\theta_i} \in \mathbb{R}^{C}$ is the combined features of classifiers parameters, instance information and semantic information, $\varphi_1$, $\varphi_2$ are linear transformations, and $\odot$ is element-wise product. Also note that $\Tilde{F}_{\theta_i}$ is further used by a classification function to produce semantic score simplex $\Delta_i$ for group classification as Equation~\ref{eq:semantic_scores}.

% To obtain the refined classifier, we learn two momentum factors $m_1, m_2 \in \mathbb{R}$ to gradually update the parameters of the classifier by:
% \begin{align}
%     \label{eq:refine_momentum}
%     m_1 = \sigma (\varphi_3(\Tilde{F}_{\theta_i})) &, m_2 = \sigma (\varphi_4(\Tilde{F}_{\theta_i})) \\
%     \Tilde{\theta}_i = m_1 \cdot \varphi_5(\Tilde{F}_{\theta_i}) &+ m_2 \cdot \varphi_6(\theta_i),
% \end{align}
% where $\sigma$ is a non-linear function sigmoid and $\varphi_3$, $\varphi_4, $\varphi_5, $\varphi_6$ are linear transformations.

\subsubsection{Classifier Self-attention}
Lastly, besides the integration of classifier parameters and their corresponding local information gathered by the attention maps, we apply self-attention to incorporate global relation into the parameters of the classifiers. We utilize a multi-head self-attention~\cite{2017Attention} to model the relation between classifiers. The relation helps classifiers distinguish between each other and understand the context of the point cloud, reducing the probability that their groupings share a large overlap and enhancing the grouping accuracy.

% The classifier refinement first highlight activated points and suppress noisy ones with the attention map $\hat{g}_i$, building on the assumption that the classifier before refinement is able to roughly distinguish between groups. Therefore, the local feature around an instance is aggregated. Moreover, the context of the classifier is modeled with self-attention. As a result, the refinement process utilize both local feature queried by the classifiers and the context information between classifiers. 
Eventually, the point features and the refined classifiers are again feed into the next stage of the decoder. The refined grouping scores generated by Equation~\ref{eq:grouping_scores} in the next stage is able to gather more points from the instance and suppress noisy ones more accurately, producing better instance feature. The refined semantic scores produced by Equation~\ref{eq:semantic_scores} assign better semantic classes for the groupings.

\input{tables/sem_test_mean}

\subsection{Context-aware CutMix}
\label{sec:cutmix}
In object detection and segmentation, class imbalance is a common issue, leading to performance degradation in the minor classes. A trivial solution to this issue is to cut objects out of training set to form a sample database. Before an input is fed into a network, the objects are sampled from the database and mix with the existing ones. 

\subsubsection{Context-aware Mixing}
We suggest mixing instances in accordance with their context in light of the aforementioned scenario.
% since recognizing instances with the understanding of its background is very useful in panoptic segmentation.
After an instance is sampled from the database, context-aware mixing translate the instance to the nearest contextual point. Contextual points are the points that most possibly exist underneath an instance, e.g., \emph{car} instances are most possibly on top of \emph{road} and \emph{parking} but not \emph{pole}. 
% We define classes of contextual points in supplementary material.
% We define classes of contextual points in~\ref{tab:contextual_points}.
% Note that pole and traffic-sign are defined as \emph{stuff} classes in SemanticKITTI, but we also apply context-aware cutmix on them since they are easy to count.
% As a result, the context information of instances is preserved.
Panoptic segmentation methods are able to model the relation between instances and background since their objective is to distinguish among them. Thus, preserving the context is beneficial to the recognition of instances.
% meanwhile the class imbalance between instances is alleviated. The grouping results of the mixed classes are enhanced.
As a result, the grouping results of the mixed classes are enhanced.
\input{tables/sem_val_mean}

%% file: tables/sem_test_mean.tex
\begin{table*}[t]
\begin{center}
% \caption{Comparison of LiDAR panoptic segmentation performance on SemanticKITTI test set. R.Net, P.P. and KPC refer to RangeNet++~\cite{RangeNet++}, Point Pillars~\cite{PointPillars} and KPConv~\cite{KPConv}, respectively. $\S$ represents results of ensemble and test-time augmentation (TTA). All scores are in [\%].}
\caption{Comparison of LiDAR panoptic segmentation performance on SemanticKITTI test set, in which PQ is the primary metric for comparison. R.Net, P.P. and KPC refer to RangeNet++~\cite{RangeNet++}, Point Pillars~\cite{PointPillars} and KPConv~\cite{KPConv}, respectively. $\S$ represents results of model ensemble and test-time augmentation (TTA). {\color[HTML]{CB0000} Red} refers to best result and {\color[HTML]{3531FF} blue} refers to second best result. Best view in color and all scores are in [\%]. Our method rank 1st on the leader borad of SemanticKITTI\footnotemark}

\label{tab:sem_test_mean}
\footnotesize

\begin{tabular}{l|
>{\columncolor[HTML]{D2FFFF}}c |ccc|ccc|ccc}
%{p{4.5cm}|p{0.5cm}p{0.5cm}p{0.5cm}|p{0.5cm}p{0.5cm}p{0.5cm}|p{0.5cm}p{0.5cm}p{0.5cm}|p{0.6cm}}
\toprule
Method & PQ & PQ$^\dagger$  & SQ & RQ & PQ\textsuperscript{Th} & SQ\textsuperscript{Th} & RQ\textsuperscript{Th} & PQ\textsuperscript{St} & SQ\textsuperscript{St} & RQ\textsuperscript{St} \\
\midrule
R.Net + P.P.      & $37.1$         & $45.9$          & $75.9$                                 & $47.0$                                 & $20.2$                                 & $75.2$                                 & $25.2$                                 & $49.3$                                 & $76.5$                                 & $62.8$                                 \\
KPC + P.P.        & $44.5$                                 & $52.5$                                 & $80.0$                                 & $54.4$                                 & $32.7$                                 & $81.5$                                 & $38.7$                                 & $53.1$                                 & $79.0$                                 & $65.9$                                 \\
Panoptic-PolarNet & $54.1$                                 & $60.7$                                 & $81.4$                                 & $65.0$                                 & $53.3$                                 & $87.2$                                 & $60.6$                                 & $54.8$                                 & $77.2$                                 & $68.1$                                 \\
DS-Net            & $55.9$                                 & $62.5$                                 & $82.3$                                 & $66.7$                                 & $55.1$                                 & $87.2$                                 & $62.8$                                 & $56.5$                                 & $78.7$                                 & $69.5$                                 \\
EfficientLPS      & $57.4$                                 & $63.2$                                 & $83.0$                                 & $68.7$                                 & $53.1$                                 & $87.8$                                 & $60.5$                                 & {\color[HTML]{3531FF} $\mathbf{60.5}$} & $79.5$                                 & {\color[HTML]{3531FF} $\mathbf{74.6}$}                                 \\
GP-S3Net          & $60.0$                                 & {\color[HTML]{3531FF} $\mathbf{69.0}$} & $82.0$                                 & $72.1$                                 & $65.0$                                 & $86.6$                                 & {\color[HTML]{CB0000} $\mathbf{74.5}$} & $56.4$                                 & $78.7$                                 & $70.4$ \\
Panoptic-PHNet    & $61.5$                                 & $67.9$                                 & {\color[HTML]{3531FF} $\mathbf{84.8}$} & $72.1$                                 & $63.8$                                 & {\color[HTML]{3531FF} $\mathbf{90.7}$} & $70.4$                                 & $59.9$                                 & {\color[HTML]{3531FF} $\mathbf{80.5}$} & $73.3$                                 \\
\midrule
\ours (ours)       & {\color[HTML]{3531FF} $\mathbf{62.2}$} & $65.8$                                 & $84.2$                                 & {\color[HTML]{3531FF} $\mathbf{72.8}$} & {\color[HTML]{3531FF} $\mathbf{65.7}$} & $90.6$                                 & $72.7$                                 & $59.6$                                 & $79.5$                                 & $73.1$                                 \\
\ours$\S$ (ours)   & {\color[HTML]{CB0000} $\mathbf{65.7}$} & {\color[HTML]{CB0000} $\mathbf{70.3}$} & {\color[HTML]{CB0000} $\mathbf{85.7}$} & {\color[HTML]{CB0000} $\mathbf{75.8}$} & {\color[HTML]{CB0000} $\mathbf{68.1}$} & {\color[HTML]{CB0000} $\mathbf{91.6}$} & {\color[HTML]{3531FF} $\mathbf{74.3}$} & {\color[HTML]{CB0000} $\mathbf{63.9}$} & {\color[HTML]{CB0000} $\mathbf{81.4}$} & {\color[HTML]{CB0000} $\mathbf{76.9}$} \\

\bottomrule
\end{tabular}
\end{center}
\end{table*}

%% file: tables/sem_val_mean.tex
\begin{table*}[t]
\begin{center}
% \caption{Comparison of LiDAR panoptic segmentation performance on SemanticKITTI validation set. R.Net, P.P. and KPC refer to RangeNet++~\cite{RangeNet++}, Point Pillars~\cite{PointPillars} and KPConv~\cite{KPConv}, respectively. $\S$ represents results of ensemble and test-time augmentation (TTA). All scores are in [\%].}
\caption{Comparison of LiDAR panoptic segmentation performance on SemanticKITTI validation set, in which PQ is the primary metric for comparison. 
R.Net, P.P. and KPC refer to RangeNet++~\cite{RangeNet++}, Point Pillars~\cite{PointPillars} and KPConv~\cite{KPConv}, respectively.
$\S$ represents results of model ensemble and test-time augmentation (TTA). {\color[HTML]{CB0000} Red} refers to best result and {\color[HTML]{3531FF} blue} refers to second best result. Best view in color and all scores are in [\%].}
\label{tab:sem_val_mean}
\footnotesize

\begin{tabular}{l|
>{\columncolor[HTML]{D2FFFF}}c |ccc|ccc|ccc}
\toprule
Method & PQ & PQ$^\dagger$  & SQ & RQ & PQ\textsuperscript{Th} & SQ\textsuperscript{Th} & RQ\textsuperscript{Th} & PQ\textsuperscript{St} & SQ\textsuperscript{St} & RQ\textsuperscript{St}  \\
\midrule

R.Net + P.P.      & $36.5$                                   &  -               & $73.0$                                 & $44.9$                                 & $19.6$                                 & $69.2$                                 & $24.9$                                 & $47.1$                                 & $75.8$                                 & $59.4$                                 \\
KPC + P.P.        & $41.1$                                   & -                                      & $74.3$                                 & $50.3$                                 & $28.9$                                 & $69.8$                                 & $33.1$                                 & $50.1$                                 & {\color[HTML]{3531FF} $\mathbf{77.6}$} & $62.8$                                 \\
DS-Net            & $57.7$                                   & $63.4$                                 & $77.6$                                 & $68.0$                                 & $61.8$                                 & $78.2$                                 & $68.8$                                 & $54.8$                                 & $77.1$                                 & $67.3$                                 \\
Panoptic-PolarNet & $59.1$                                   & $64.1$                                 & $78.3$                                 & $70.2$                                 & $65.7$                                 & $87.4$                                 & $74.7$                                 & $54.3$                                 & $71.6$                                 & $66.9$                                 \\
EfficientLPS      & $59.2$                                   & $65.1$                                 & $75.0$                                 & $69.8$                                 & $58.0$                                 & $78.0$                                 & $68.2$                                 & {\color[HTML]{CB0000} $\mathbf{60.9}$} & $72.8$                                 & $71.0$                                 \\
Panoptic-PHNet    & $61.7$                                   & -                                      & -                                      & -                                      & $69.3$                                 & -                                      & -                                      & -                                      & -                                      & -                                 \\
GP-S3Net          & $63.3$                                   & $67.5$                                 & $81.4$                                 & {\color[HTML]{CB0000} $\mathbf{75.9}$} & $70.2$                                 & $86.2$                                 & {\color[HTML]{3531FF} $\mathbf{80.1}$} & $58.3$                                 & {\color[HTML]{CB0000} $\mathbf{77.9}$} & {\color[HTML]{3531FF} $\mathbf{71.9}$} \\
\midrule
\ours (ours)       & {\color[HTML]{3531FF} $\mathbf{64.4}$} & {\color[HTML]{3531FF} $\mathbf{68.6}$} & {\color[HTML]{3531FF} $\mathbf{81.5}$} & $74.1$                                 & {\color[HTML]{3531FF} $\mathbf{73.0}$} & {\color[HTML]{3531FF} $\mathbf{92.6}$} & $79.3$                                 & $58.1$                                 & $73.5$                                 & $70.4$                                 \\
\ours$\S$ (ours)   & {\color[HTML]{CB0000} $\mathbf{66.3}$}   & {\color[HTML]{CB0000} $\mathbf{70.2}$} & {\color[HTML]{CB0000} $\mathbf{82.5}$} & {\color[HTML]{3531FF} $\mathbf{75.6}$} & {\color[HTML]{CB0000} $\mathbf{74.6}$} & {\color[HTML]{CB0000} $\mathbf{93.4}$} & {\color[HTML]{CB0000} $\mathbf{80.3}$} & {\color[HTML]{3531FF} $\mathbf{60.2}$} & $74.5$                                 & {\color[HTML]{CB0000} $\mathbf{72.2}$} \\          

\bottomrule
\end{tabular}
\end{center}
\end{table*}

%% file: sections/experiment.tex
\section{Experiment}
% \label{sec:exp}
%         \input{tables/sem_test_mean}
    To evaluate \ours, we conduct experiments on two popular LiDAR point cloud datasets: SemanticKITTI\cite{SemanticKITTI} and nuScenes\cite{nuScenesSeg}.
    
    \subsection{Datasets and Evaluation Metric}
    \label{sec:dataset_metric}
        \subsubsection{SemanticKITTI} \label{sec:semantickitti}
        proposes the first panoptic segmentation challenge on point cloud data. It contains 22 data sequences splited into 3 parts: 10 for training, 1 for validation and 11 for testing. There are 8 \emph{thing} classes and 11 \emph{stuff} classes.
        
        \subsubsection{nuScenes} \label{sec:nuscenes}
        is a large-scale dataset for autonomous driving, which contains LiDAR data of 1000 scenes. The 1000 scenes are divided into 3 parts: 750 for training, 100 scenes for validation and 150 scenes for testing. There are 10 \emph{thing} classes and 6 \emph{stuff} classes.
        
        \subsubsection{Evaluation Metric} \label{sec:metric}
        Mean Panoptic Quality (PQ) \cite{panopticsegmentation} is adopted as the primary evaluation metric for our experiment. As shown in Equation \ref{eq:pqc}, PQ of a specific class can be decomposed into Segmentation Quality (SQ) and Recognition Quality (RQ):
        
        \begin{equation}
            \label{eq:pqc}
            \textrm{PQ}_c = \underbrace{\frac{\sum_{(p, g)\in \textrm{TP}_c}  \textrm{IoU}(p, g) }{|\textrm{TP}_c|}}_{\textrm{segmentation quality (SQ)}} 
            \times \underbrace{\frac{|\textrm{TP}_c|}{|\textrm{TP}_c| + \frac{1}{2} |\textrm{FP}_c| + \frac{1}{2} |\textrm{FN}_c|}}_{\textrm{recognition quality (RQ)} },
        \end{equation}
        where $\textrm{TP}_c$ is the set of matched predicted masks and ground truth masks of class $c$, $\textrm{FP}_c$ is the set of unmatched predicted masks of class $c$, $\textrm{FN}_c$ is set of unmatched ground truth masks of class c, and $\textrm{IoU}(p, g)$ is the intersection-over-union of predicted mask $p$ and ground truth mask $g$. 
        % It is worth noting that for each pair of masks $(p, g)$ in $\textrm{TP}_c$, not only $p$ and $g$ are of the same class $c$, but also their $\textrm{IoU}(p, g)$ exceed a threshold of 0.5.
        % Mean PQ is the average of PQ of all classes as shown in Equation \ref{eq:pq}:
        % \begin{equation}
        %     \label{eq:pq}
        %     \textrm{PQ} = \frac{1}{T} \sum_{i=1}^{T} \textrm{PQ}_i,
        % \end{equation}
        % where $T$ number of \emph{thing} and \emph{stuff} classes. Similar to Equation \ref{eq:pqc} and Equation \ref{eq:pq}, we additionally report $\textrm{PQ}^{\textrm{Th}}$, $\textrm{SQ}^{\textrm{Th}}$ and $\textrm{RQ}^{\textrm{Th}}$ of \emph{thing} classes, $\textrm{PQ}^{\textrm{St}}$, $\textrm{SQ}^{\textrm{St}}$ and $\textrm{RQ}^{\textrm{St}}$ of \emph{stuff} classes and $\textrm{PQ}^{\dagger}$ \cite{pqdagger}, where $\textrm{PQ}$ of \emph{stuff} classes is replaced by their $\textrm{IoU}$ in the calculation.
        Mean PQ is the average of PQ of all classes and we additionally report $\textrm{PQ}^{\textrm{Th}}$, $\textrm{SQ}^{\textrm{Th}}$ and $\textrm{RQ}^{\textrm{Th}}$ of \emph{thing} classes, $\textrm{PQ}^{\textrm{St}}$, $\textrm{SQ}^{\textrm{St}}$ and $\textrm{RQ}^{\textrm{St}}$ of \emph{stuff} classes and $\textrm{PQ}^{\dagger}$ \cite{pqdagger}, where $\textrm{PQ}$ of \emph{stuff} classes is replaced by their $\textrm{IoU}$ in the calculation.

    \subsection{Implementation Details}
    \label{sec:implementation_details}
    \subsubsection{Settings and Hyper-parameters}
    Our implementation is based on MMDetection3D~\cite{mmdet3d2020}. Specifically, we train our models for 80 epochs with a batch size of 4. The learning rate is set to 0.002 initially and decrease with a factor of 0.1 after 50 epochs. We adopt AdamW~\cite{adamw} with a weight decay of 0.05 as our optimizer. In addition to our proposed CutMix augmentation, we apply random flipping along x- and y- axis, random rotation along z- axis and random scaling. Unless specified, the point feature dimension is set to 128 and the number of classifiers is set to 100. The number of refinement stages is 3. As for training, the losses are included in Equation~\ref{eq:match_loss} and the coefficients $\alpha$, $\beta$, $\gamma$ are set to 4, 1, 1 respectively. 
    % Grouping scores and semantic scores from all stages contribute to training loss.
    
        \footnotetext{Our 1st place performance is assessed on Aug 12th, 2022 in \emph{https://competitions.codalab.org/competitions/24025\#results}.}
    \subsubsection{Backbone}
    We employ the backbone of RPVNet~\cite{RPVNet} in \ours. It fuse points, voxels and range-view feature, and extract representative features. We follow the same backbone architecture as RPVNet but change the output feature dimension to 128. The voxel size is set to 5 cm and 10 cm for SemanticKITTI and nuScenes respectively.
        \input{tables/nus_val_mean}
    
        \input{tables/ablation_num_stages}
    \input{tables/ablation_components}
    \subsection{Main Results}
    \label{sec:main_results}
        \subsubsection{Results on SemanticKITTI} 
        \label{sec:result_sem}
        % In this section, we evaluate \ours on SemanticKITTI and compare its panoptic segmentation results with state-of-the-art methods.
        As shown in Table~\ref{tab:sem_test_mean} and \ref{tab:sem_val_mean}, we surpass all existing methods in PQ of both test set and validation set, and show significant advantages in the performance of \emph{thing} classes. As for test set, we improve PQ of Panoptic-PHNet~\cite{Panoptic_PHNet} from $61.5\%$ to $62.2\%$ and achieve a gain of $1.9\%$ in $\textrm{PQ}^{\textrm{Th}}$. As for validation set, we outperform GP-S3Net~\cite{GP-S3Net} by a margin of $1.1\%$ in PQ and $2.8\%$ in $\textrm{PQ}^{\textrm{Th}}$.
        Compared with clustering-based methods DS-Net~\cite{DS_Net}, and Panoptic-PolarNet~\cite{Panoptic-PolarNet} in addition to Panoptic-PHNet and GP-S3Net, our method achieve an increase of over $6\%$ in PQ of test set. With respect to range-image-based method EfficientLPS~\cite{EfficientLPS}, \ours outperforms by $4.8\%$ in PQ of test set. The results of combined methods (row 1 and row 2) presented in the tables are obtained by training a detection head and semantic head as stated in the dataset. 
        Moreover, following Panoptic-PHNet, we report results of model ensemble and test-time augmentation.
        Additionally, we provide class-wise performance of \ours in supplementary material.
        
        \subsubsection{Results on nuScenes} 
        \label{sec:result_nus}
        In this section, we compare the results of \ours on nuScenes with results of previous methods. As listed in Table~\ref{tab:nus_val_mean}, our method achieves state-of-the-art results on validation set.
        
    \subsection{Ablation Study}
    % To analyze the effectiveness of our network components, we conduct ablation study based on validation set of SemanticKITTI.
    
        \subsubsection{Ablation on Network Components}
        To verify the effectiveness of \ours, we gradually apply our proposed components to a vanilla network. As shown in Table~\ref{tab:ablation_components}, M1 refers to a vanilla network with no refinement on the classifier or CutMix augmentation. M2 is trained with classifier refinement and M3 is trained with context-aware CutMix. The performance of M4 shows that both classifier refinement and context-aware CutMix contribute to the high performance.

        \subsubsection{Ablation on Number of Stages} As shown in Table~\ref{tab:ablation_num_stages}, trials with different number of stages reveals that \ours with 3 stages achieves the best result. We observe that there exists over-fitting concerning the decrease of training loss as the number of stages increase. It suggests that models for larger datasets may benefit from more stages.
        
        \subsubsection{Ablation on Number of Classifiers} Table~\ref{tab:ablation_num_classifiers} contains result from different number of classifiers. It shows that 100 classifiers achieve the best result. On the one hand, insufficient number of classifiers is harmful to performance on both \emph{thing} classes and \emph{stuff} classes since bipartite assignment may assign the classifiers to inconsistent semantic. On the other hand, excessive number of classifiers benefit from consistency in assignment and perform better in \emph{thing} classes. 
        % However, background classes are over segmented by large number of classifiers, leading to a drop in \emph{stuff} classes.
        However, since the number of classifiers for background classes is fixed, excessive instance classifiers may lead to under-segmented background.
        
        % \subsubsection{Ablation on Voxel Size} As for voxel sizes used in our backbone, we adopt cubic voxel and set its dimension to different sizes. Tab~\ref{tab:ablation_voxel_size} shows that voxel size of 5 cm is better since large voxel size leads to large voxelization error when reducing point features to voxel features and small voxel size results in isolation of voxels in space during sparse operations like sparse convolution.
        % \input{tables/ablation_voxel_size}
        
        % \subsubsection{Ablation on Encoder Feature Dimension}
        % \input{tables/ablation_feature_dim}
        
        \input{tables/ablation_num_classifiers}
        
        \subsubsection{Ablation on CutMix Strategies} In addition to our proposed context-aware CutMix, there is another CutMix strategy in point cloud object detection and segmentation: random CutMix~\cite{SECOND, RPVNet, Panoptic_PHNet}. They alleviate class imbalance by randomly mixing the sampled instances in the current scan. To validate the effectiveness of our context-aware CutMix, we compare performance by applying the strategies on M1 in Table~\ref{tab:ablation_strategy}. As shown in Table~\ref{tab:ablation_strategy}, our context-aware CutMix achieve a gain of 5.1\% in PQ and outperform by a large margin on \emph{thing} classes. It verifies our design on preserving context information of instances to enhance performance.
        
        \input{tables/ablation_strategy}
        
    \subsection{Analysis and Visualization}
    
        \subsubsection{Spatial Distributions of Predictions}
\begin{figure}[h]
            \centering
            % \vspace{-7mm}
            % \includegraphics[width=0.47\textwidth]{figures/spatial_locations_alpha1.pdf}
            % \includegraphics[width=0.3\textwidth]{figures/spatial_locations_alpha1.pdf}
            \includegraphics[width=0.45\textwidth]{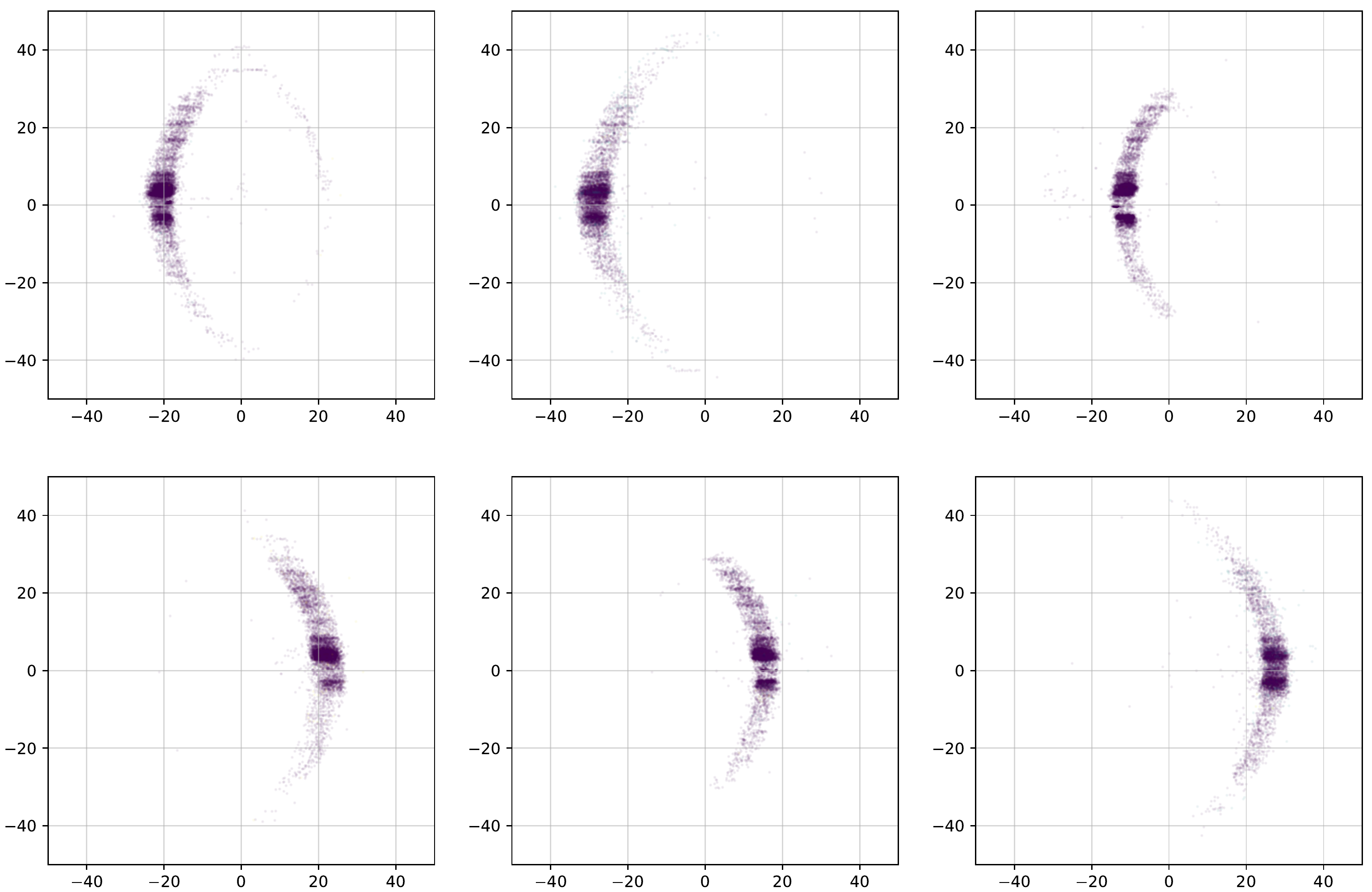}
            % \vspace{-3mm}
            \caption{Spatial distributions of predictions by classifiers. The centers are projected into a 100m $\times$ 100m x-y plane. Results are obtained from SemanticKITTI test set.}
            \label{fig:spatial}
        \end{figure}

        As stated in Section~\ref{sec:bipartite_matching} and ~\ref{sec:classifier_refine}, our classifiers are able to distinguish between instances, hence being capable of predicting panoptic segmentation results directly. Considering that the predictions are in 3D space, it is better to present an illustration more intuitively. 
        Therefore, we plot the centers of the predictions in bird-eye view (BEV). As shown in Figure~\ref{fig:spatial}, each subplot stands for the spatial distribution of a classifier's predictions on \emph{car} in a 100m $\times$ 100m square. The distributions follow certain patterns: 1) the arc-shaped patterns reveal accordance with the rotation of LiDAR sensors. 2) the positions where dense predictions located demonstrate spatial spacing, verifying the ability of the classifiers to predict exclusive instances.
        % \begin{figure}
        %     \centering
        %     % \includegraphics[width=0.47\textwidth]{figures/spatial_locations.pdf}
        %     \includegraphics[width=0.47\textwidth]{figures/spatial_locations_alpha0.1.pdf}
        %     \caption{Spatial distributions of predictions by classifiers.}
        %     % \label{fig:intropic}
        % \end{figure}
        
        % \input{tables/sem_test_classwise}
        
        % \input{tables/nus_test_mean}

%% file: tables/nus_val_mean.tex
\begin{table*}[t!]
\begin{center}
\caption{Comparison of LiDAR panoptic segmentation performance on nuScenes validation set. {\color[HTML]{CB0000} Red} refers to best result and {\color[HTML]{3531FF} blue} refers to second best result. Best view in color and all scores are in [\%].}
\label{tab:nus_val_mean}
\footnotesize

\begin{tabular}{l|
>{\columncolor[HTML]{D2FFFF}}c| ccc|ccc|ccc}
\toprule
Method & PQ & PQ$^\dagger$  & SQ & RQ & PQ\textsuperscript{Th} & SQ\textsuperscript{Th} & RQ\textsuperscript{Th} & PQ\textsuperscript{St} & SQ\textsuperscript{St} & RQ\textsuperscript{St} \\
\midrule
PanopticTrackNet  & $51.4$                                 & $56.2$                                 & $80.2$                                 & $63.3$                                 & $45.8$                                 & $81.4$                                 & $55.9$                                 & $60.4$                                 & $78.3$                                 & $75.5$                                 \\
DS-Net                    & $55.9$ & $62.5$ & $82.3$ & $66.7$ & $55.1$ & $87.2$ & $62.8$ & $56.5$ & $78.7$ & $69.5$ \\
GP-S3Net          & $61.0$                                 & $67.5$                                 & $84.1$                                 & $72.0$                                 & $56.0$                                 & $85.3$                                 & $65.2$                                 & $66.0$                                 & $82.9$                                 & $78.7$                                 \\
EfficientLPS      & $62.0$                                 & $65.6$                                 & $83.4$                                 & $73.9$                                 & $56.8$                                 & $83.2$                                 & $68.0$                                 & $70.6$                                 & $83.8$                                 & $83.6$                                 \\
Panoptic-PolarNet & $63.4$                                 & $67.2$                                 & $83.9$                                 & $75.3$                                 & $59.2$                                 & $84.1$                                 & $70.3$                                 & $70.4$                                 & $83.6$                                 & $83.5$                                 \\
Panoptic-PHNet    & {\color[HTML]{CB0000} $\mathbf{74.7}$} & {\color[HTML]{CB0000} $\mathbf{77.7}$} & {\color[HTML]{3531FF} $\mathbf{88.2}$} & {\color[HTML]{CB0000} $\mathbf{84.2}$} & {\color[HTML]{CB0000} $\mathbf{74.0}$} & {\color[HTML]{3531FF} $\mathbf{89.0}$} & {\color[HTML]{CB0000} $\mathbf{82.5}$} & {\color[HTML]{CB0000} $\mathbf{75.9}$} & {\color[HTML]{CB0000} $\mathbf{86.8}$} & {\color[HTML]{CB0000} $\mathbf{86.9}$} \\
\midrule
% ours (ours)       & {\color[HTML]{3531FF} $\mathbf{71.0}$} & {\color[HTML]{3531FF} $\mathbf{73.8}$} & {\color[HTML]{CB0000} $\mathbf{88.3}$} & {\color[HTML]{3531FF} $\mathbf{80.0}$} & {\color[HTML]{3531FF} $\mathbf{69.9}$} & {\color[HTML]{CB0000} $\mathbf{90.4}$} & {\color[HTML]{3531FF} $\mathbf{77.0}$} & {\color[HTML]{3531FF} $\mathbf{72.7}$} & {\color[HTML]{3531FF} $\mathbf{84.7}$} & {\color[HTML]{3531FF} $\mathbf{85.2}$} \\
% ours (ours)       & {\color[HTML]{3531FF} $\mathbf{72.8}$} & {\color[HTML]{3531FF} $\mathbf{75.6}$} & {\color[HTML]{CB0000} $\mathbf{89.1}$} & {\color[HTML]{3531FF} $\mathbf{81.4}$} & {\color[HTML]{3531FF} $\mathbf{72.6}$} & {\color[HTML]{CB0000} $\mathbf{91.5}$} & {\color[HTML]{3531FF} $\mathbf{79.1}$} & {\color[HTML]{3531FF} $\mathbf{73.2}$} & {\color[HTML]{3531FF} $\mathbf{85.2}$} & {\color[HTML]{3531FF} $\mathbf{85.2}$} \\
% ours (ours)       & {\color[HTML]{3531FF} $\mathbf{74.4}$} & {\color[HTML]{3531FF} $\mathbf{77.0}$} & {\color[HTML]{CB0000} $\mathbf{89.3}$} & {\color[HTML]{3531FF} $\mathbf{83.0}$} & {\color[HTML]{3531FF} $\mathbf{74.9}$} & {\color[HTML]{CB0000} $\mathbf{91.8}$} & {\color[HTML]{3531FF} $\mathbf{81.4}$} & {\color[HTML]{3531FF} $\mathbf{73.6}$} & {\color[HTML]{3531FF} $\mathbf{85.3}$} & {\color[HTML]{3531FF} $\mathbf{85.6}$} \\
\ours (ours)       & {\color[HTML]{CB0000} $\mathbf{74.7}$} & {\color[HTML]{3531FF} $\mathbf{77.3}$} & {\color[HTML]{CB0000} $\mathbf{89.4}$} & {\color[HTML]{3531FF} $\mathbf{83.3}$} & {\color[HTML]{3531FF} $\mathbf{75.4}$} & {\color[HTML]{CB0000} $\mathbf{91.8}$} & {\color[HTML]{3531FF} $\mathbf{81.9}$} & {\color[HTML]{3531FF} $\mathbf{73.6}$} & {\color[HTML]{3531FF} $\mathbf{85.3}$} & {\color[HTML]{3531FF} $\mathbf{85.6}$} \\
\bottomrule
\end{tabular}
\end{center}
\end{table*}

%% file: tables/ablation_num_stages.tex
\begin{table}[ht]\small
\begin{center}
\caption{Ablation study of number of refining stages on validation set of SemanticKITTI. $\mathcal{L}_{\text{train}}$ denotes training loss of models. All scores are in [\%]. }
\setlength{\tabcolsep}{2.4mm}
\centering
\label{tab:ablation_num_stages}
\begin{tabular}{c|ccc|c}
\toprule
\# of Stages    & PQ              & SQ              & RQ              & $\mathcal{L}_{\text{train}}$ \\
\midrule
1               & $62.1$            & $80.6$            & $72.0$            & $0.213$\\
2               & $63.5$            & $81.0$            & $73.2$            & $0.139$\\
3               & $\mathbf{64.4}$   & $\mathbf{81.5}$   & $\mathbf{74.1}$   & $0.125$\\
4               & $63.2$            & $80.8$            & $72.9$            & $0.113$\\
5               & $63.0$            & $80.5$            & $73.0$            & $0.101$\\
\bottomrule
\end{tabular}
\end{center}
\end{table}

%% file: tables/ablation_components.tex
% \begin{table*}[htbp]
% \setlength\tabcolsep{3.5pt}
% \centering
% \caption{Ablation study on proposed components of \ours. The results are reported on the SemanticKITTI validation set.} 
% \label{tab:ablation_components}
% \begin{tabular}{@{}c|ccc|cccc|ccc|ccc|c}
% \toprule
% Model & Random & Contextual & Classifier & PQ & PQ$^\dagger$ & SQ & RQ & PQ\textsuperscript{Th} & SQ\textsuperscript{Th} & RQ\textsuperscript{Th} & PQ\textsuperscript{St} & SQ\textsuperscript{St} & RQ\textsuperscript{St} & mIoU\\
% Variant & Cutmix & CutMix & Refinement & (\%) & (\%) & (\%) & (\%) & (\%) & (\%) & (\%) & (\%) & (\%) & (\%) & (\%)\\
% \midrule
% M1 & \, & \, & \,     & $44.4$ & $49.4$ & $68.7$ & $53.8$ & $37.8$ & $74.7$ & $43.9$ & $49.1$ & $64.4$ & $61.1$ & $46.7$ \\ 
% M2 & \cmark & \, & \, & $50.8$ & $56.3$ & $72.8$ & $61.7$ & $57.5$ & $86.4$ & $66.4$ & $50.0$ & $62.9$ & $58.2$ & $56.4$ \\
% M3 & \, & \cmark & \, & $55.9$ & $60.7$ & $75.1$ & $66.2$ & $64.4$ & $90.5$ & $71.8$ & $49.7$& $63.9$& $62.1$ & $62.2$ \\
% M4 & \, & \cmark & \cmark & $\mathbf{64.4}$& $\mathbf{68.6}$ & $\mathbf{81.5}$ & $\mathbf{74.1}$ & $\mathbf{73.0}$ & $\mathbf{92.6}$& $\mathbf{79.3}$ & $\mathbf{58.1}$ &$\mathbf{73.5}$ & $\mathbf{70.4}$ & $\mathbf{66.9}$ \\
% \bottomrule
% \end{tabular}
% \end{table*}

\begin{table*}[t]
\setlength\tabcolsep{3.5pt}
\centering
\caption{Ablation study on proposed components of \ours. The results are reported on the SemanticKITTI validation set.} 
\label{tab:ablation_components}
\begin{tabular}{@{}c|cc|cccc|ccc|ccc}
\toprule
Model & Context-aware & Classifier & PQ & PQ$^\dagger$ & SQ & RQ & PQ\textsuperscript{Th} & SQ\textsuperscript{Th} & RQ\textsuperscript{Th} & PQ\textsuperscript{St} & SQ\textsuperscript{St} & RQ\textsuperscript{St} \\
Variant & CutMix & Refinement & (\%) & (\%) & (\%) & (\%) & (\%) & (\%) & (\%) & (\%) & (\%) & (\%) \\
\midrule
M1 & \, & \,     & $44.5$ & $49.3$ & $68.0$ & $54.4$ & $40.5$ & $74.6$ & $47.0$ & $47.4$ & $63.2$ & $59.7$  \\ 
M2 & \, & \cmark & $50.9$ & $55.1$ & $74.2$ & $60.5$ & $43.6$ & $76.5$ & $49.5$ & $56.2$ & $72.5$ & $56.2$  \\
M3 & \cmark & \, & $55.9$ & $60.7$ & $75.1$ & $66.2$ & $64.4$ & $90.5$ & $71.8$ & $49.7$& $63.9$& $62.1$ \\
M4 & \cmark & \cmark & $\mathbf{64.4}$& $\mathbf{68.6}$ & $\mathbf{81.5}$ & $\mathbf{74.1}$ & $\mathbf{73.0}$ & $\mathbf{92.6}$& $\mathbf{79.3}$ & $\mathbf{58.1}$ &$\mathbf{73.5}$ & $\mathbf{70.4}$  \\
\bottomrule
\end{tabular}
\end{table*}

%% file: tables/ablation_num_classifiers.tex
\begin{table}[htbp]\small
\begin{center}
\caption{Ablation study of number of classifiers on validation set of SemanticKITTI. All scores are in [\%].}
\setlength{\tabcolsep}{2mm}
\centering
\label{tab:ablation_num_classifiers}
\begin{tabular}{c|ccc|c|c}
\toprule
\# of Classifiers& PQ              & SQ              & RQ              & PQ\textsuperscript{Th} & PQ\textsuperscript{St} \\
\midrule
50               & $63.3$          & $80.0$          & $73.0$          & $69.8$                 & $56.2$                   \\
100              & $\mathbf{64.4}$ & $\mathbf{81.5}$ & $\mathbf{74.1}$ & $73.0$                 & $\mathbf{58.1}$        \\
150              & $63.7$          & $81.0$          & $73.6$          & $73.7$                 & $56.3$                   \\
200              & $63.8$          & $81.0$          & $73.9$          & $\mathbf{73.9}$        & $56.5$                   \\

\bottomrule
\end{tabular}
\end{center}
\end{table}

%% file: tables/ablation_strategy.tex
\begin{table}[htbp]\small
\begin{center}
\caption{Ablation study of cutmix strategy on validation set of SemanticKITTI. All scores are in [\%].}
\setlength{\tabcolsep}{1.2mm}
\centering
\label{tab:ablation_strategy}
\begin{tabular}{c|ccc|ccc}
\toprule
CutMix Type      & PQ              & SQ              & RQ              & PQ\textsuperscript{Th} & SQ\textsuperscript{Th} & RQ\textsuperscript{Th} \\
\midrule
Random           & $50.8$          & $72.8$          & $61.7$          & $57.5$                 & $86.4$                 & $66.4$                   \\
Context-aware    & $\mathbf{55.9}$ & $\mathbf{75.1}$ & $\mathbf{66.2}$ & $\mathbf{64.4}$        & $\mathbf{90.5}$        & $\mathbf{71.8}$        \\

\bottomrule
\end{tabular}
\end{center}
\end{table}

%% file: sections/conclusion.tex
% \section{Conclusion}
% In this paper, we develop a unified panoptic segmentation framework for point cloud data, which is capable to exclusively predict panoptic results without any hand-crafted post-processing and achieves state-of-the-art performance. \ours allocates a set of classifiers to learn to group coherent points in a point cloud and introduce bipartite matching into a point cloud network to enable end-to-end training. Moreover, \ours employs a transformer decoder to refine the groupings and resolve class imbalance in point cloud panoptic segmentation by designing a context-aware cutmix augmentation. \ours is the first to provide a holistic and end-to-end solution to point cloud panoptic segmentation. 
% We hope that \ours can inspire more researchers to delve into the development of unified segmentation for point cloud, which is beneficial to promoting autonomous vehicles.

\section{Conclusion}
In this paper, we develop a unified panoptic segmentation framework, dubbed \ours, for point cloud data, which is capable to exclusively predict panoptic results without any hand-crafted post-processing and achieves state-of-the-art performance. \ours allocates a set of classifiers to learn how to group coherent points directly and introduces bipartite matching to enable end-to-end training. Moreover, \ours employs a transformer decoder to refine the groupings and resolve class imbalance problem by designing a context-aware cutmix augmentation. \ours is the first to provide a holistic and end-to-end solution for point cloud panoptic segmentation. 
We hope that \ours can inspire more researchers to delve into the development of unified segmentation for point cloud, which is beneficial to promoting autonomous vehicles.